\documentstyle[11pt,a4]{article}
\begin{document}

\begin{center}
{\Large {\bf Prepositional Phrase Attachment through a Backed-Off Model}}
\end{center}

\begin{center}
{\bf Michael Collins} and {\bf James Brooks}

Department of Computer and Information Science

University of Pennsylvania

Philadelphia, PA 19104

\{mcollins, jbrooks\}@gradient.cis.upenn.edu
\end{center}

\setlength{\parindent}{0in}
\setlength{\parskip}{2ex}

\begin{center}
{\bf Abstract}
\end{center}

Recent work has considered corpus-based or statistical approaches to
the problem of prepositional phrase attachment ambiguity.
Typically, ambiguous verb phrases of the form {\it v np1 p np2} are
resolved through a model which considers values of the four head
words ($v$, $n1$, $p$ and $n2$). This paper shows that the
problem is analogous to n-gram language models in speech
recognition, and that one of the most common methods for language modeling,
the backed-off estimate, is applicable. Results on Wall
Street Journal data of 84.5\% accuracy are obtained using this method.
A surprising result is the importance of
low-count events - ignoring events which occur less than 5
times in training data reduces performance to 81.6\%.

\section{Introduction}

Prepositional phrase attachment is a common cause of
structural ambiguity in natural language. For example take the
following sentence:

\begin{center}
Pierre Vinken, 61 years old, {\em joined the board as a
nonexecutive director}.
\end{center}

The PP `as a nonexecutive director' can either attach to the NP `the
board' or to the VP `joined', giving two alternative structures.
(In this case the VP attachment is correct):

\begin{center}
NP-attach: (joined ((the board) (as a nonexecutive director)))
\end{center}

\begin{center}
VP-attach: ((joined (the board)) (as a nonexecutive director))
\end{center}

Work by Ratnaparkhi, Reynar and Roukos \cite{rrr}
and Brill and Resnik \cite{br}
has considered corpus-based approaches
to this problem, using a set of examples to
train a model which is then
used to make attachment decisions on test data. Both papers
describe methods which look at the four head words involved in the
attachment - the VP head, the first NP head, the preposition and the second
NP head (in this case {\em joined}, {\em board}, {\em as} and {\em
director} respectively).

This paper proposes a new statistical method for PP-attachment
disambiguation based on the four head words.

\section{Background}

\subsection{Training and Test Data}

The training and test data were supplied by IBM, being identical to
that used in \cite{rrr}. Examples of verb phrases
containing a (v np pp) sequence had been taken from the Wall Street
Journal Treebank \cite{msm}.
For each such VP the head verb, first head noun, preposition
and second head noun were extracted, along with the attachment
decision (1 for noun attachment, 0 for verb). For example the
verb phrase:

\begin{center}
((joined (the board)) (as a nonexecutive director))
\end{center}

would give the quintuple:

\begin{center}
0 joined board as director
\end{center}

The elements of this quintuple will from here on be referred to
as the random variables $A$, $V$, $N1$, $P$, and $N2$. In the above
verb phrase $A=0$, $V=joined$, $N1=board$, $P=as$, and $N2=director$.

The data consisted of training and test files of 20801 and 3097 quintuples
respectively. In addition, a development set of
4039 quintuples was also supplied. This set was used during development
of the attachment algorithm, ensuring that there was no
implicit training of the method on the test set itself.

\subsection{Outline of the Problem}

A PP-attachment algorithm must take each quadruple
($V=v$, $N1=n1$, $P=p$, $N2=n2$) in test data and decide whether
the attachment variable $A$ = 0 or 1. The accuracy
of the algorithm is then the percentage of attachments
it gets `correct' on test data, using the $A$ values taken from the treebank
as the reference set.

The probability of the attachment variable $A$
being 1 or 0 (signifying noun or verb attachment respectively) is
a probability, $p$, which is conditional on the values of the words
in the quadruple. In general a probabilistic algorithm will make an
estimate, $\hat{p}$, of this probability:

\[\hat{p}(A=1|V=v,N1=n1,P=p,N2=n2)\]

For brevity this estimate will be referred to from here on as:

\[\hat{p}(1|v,n1,p,n2)\]

The decision can then be made using the test:

\[\hat{p}(1|v,n1,p,n2)>=0.5\]

If this is true the attachment is made to the noun, if not then
it is made to the verb.

\subsection{Lower and Upper Bounds on Performance}

When evaluating an algorithm it is useful to have an
idea of the lower and upper bounds on its performance.
Some key results are summarised in the table below. All
results in this section are on the IBM training and test data,
with the exception of the two `average human' results.

\begin{center}\begin{tabular}{|c|c|}\hline
Method&Percentage Accuracy\\ \hline \hline
Always noun attachment&59.0\\ \hline
Most likely for each preposition&72.2\\ \hline
Average Human (4 head words only)&88.2\\ \hline
Average Human (whole sentence)&93.2\\ \hline
\end{tabular}\end{center}

`Always noun attachment' means attach to the noun regardless of
(v,n1,p,n2). `Most likely for each preposition' means use the
attachment seen most often in training data for the
preposition seen in the test quadruple. The
human performance results are taken from \cite{rrr}, and are
the average performance of 3 treebanking experts on a set of 300
randomly selected test events from the WSJ corpus, first looking at
the four head words alone, then using the whole sentence.

A reasonable lower bound seems to be 72.2\% as scored by the
`Most likely for each preposition' method. An approximate upper bound
is 88.2\% - it seems unreasonable to expect an algorithm to perform
much better than a human.

\section{Estimation based on Training Data Counts}

\subsection{Notation}

We will use the symbol $f$ to denote the number of times a
particular tuple is seen in training data. For example
$f(1,is,revenue,from,research)$ is the number of times the quadruple
$(is,revenue,from,research)$ is seen with a noun attachment. Counts of
lower order tuples can also be made - for example $f(1,P=from)$
is the number of times $(P=from)$ is seen with noun attachment in
training data, $f(V=is,N2=research)$ is the number of times
$(V=is,N2=research)$ is seen with either attachment and any value of N1
and P.

\subsection{Maximum Likelihood Estimation}

A maximum likelihood method would use the training data to give the
following estimation for the conditional probability:

\[\hat{p}(1|v,n1,p,n2)=\frac{f(1,v,n1,p,n2)}{f(v,n1,p,n2)}\]

Unfortunately sparse data problems make this estimate useless.
A quadruple may appear in test data which has never been seen in
training data. ie. $f(v,n1,p,n2)=0$. The above estimate is undefined
in this situation, which happens extremely frequently in a large
vocabulary domain such as WSJ. (In this experiment about 95\% of those
quadruples appearing in test data had not been seen in training data).

Even if $f(v,n1,p,n2)>0$, it may still be very low, and this may make
the above MLE estimate inaccurate. Unsmoothed MLE estimates based on
low counts are notoriously bad in similar problems such as n-gram language
modeling \cite{gc}. However later in this paper it is shown that
estimates based on low counts are surprisingly useful in the
PP-attachment problem.

\subsection{Previous Work}

Hindle and Rooth \cite{hr} describe one of the first statistical approaches to
the prepositional phrase attachment problem. Over 200,000
$(v,n1,p)$ triples were extracted from 13 million words of AP news
stories. The attachment decisions for these triples were unknown, so
an unsupervised training method was used (section 5.2 describes the
algorithm in more detail). Two human judges annotated
the attachment decision for 880 test examples, and the method
performed at 80\% accuracy on these cases. Note that it is difficult
to compare this result to results on Wall Street Journal, as the two
corpora may be quite different.

The Wall Street Journal Treebank \cite{msm} enabled both \cite{rrr} and
\cite{br} to extract a large amount of supervised training material for the
problem. Both of these methods consider the second noun,
 $n2$, as well as $v$, $n1$ and $p$, with the hope that this additional
information will improve results.

\cite{br} use 12,000 training and 500 test examples. A greedy search is
used to learn a sequence of `transformations' which minimise the error
rate on training data. A transformation is a rule which makes an
attachment decision depending on up to 3 elements of the $(v,n1,p,n2)$
quadruple. (Typical examples would be `If P={\it
of} then choose noun attachment' or `If V={\it buy} and P={\it for}
choose verb attachment').
A further experiment incorporated word-class information from WordNet
into the model, by allowing the transformations to look at classes as
well as the words. (An example would be `If N2 is in
the {\it time} semantic class, choose verb attachment'). The method
gave 80.8\% accuracy with words only, 81.8\% with words and semantic classes,
and they also report an accuracy of 75.8\% for the metric of \cite{hr} on this
data. Transformations (using words only) score 81.9\%\footnote{Personal
communication from Brill.} on the IBM data used in this paper.

\cite{rrr} use the data described in section 2.1 of this paper - 20801
training and 3097 test examples from Wall Street Journal. They use a
maximum entropy model which also considers subsets of the
quadruple. Each sub-tuple predicts noun or verb attachment with
a weight indicating its strength of prediction - the weights are
trained to maximise the likelihood of training data. For example
$(P=of)$ might have a strong weight for noun attachment, while
$(V=buy,P=for)$ would have a strong weight for verb attachment. \cite{rrr}
also allow the model to look at class information, this time the
classes were learned automatically from a corpus. Results of 77.7\%
(words only) and 81.6\% (words and classes) are reported. Crucially they
ignore low-count events in training data by imposing a
frequency cut-off somewhere between 3 and 5.

\section{The Backed-Off Estimate}

\cite{k} describes backed-off n-gram word models for speech
recognition. There the task is to estimate the probability of the next word
in a text given the (n-1) preceding words. The MLE estimate
of this probability would be:

\[\hat{p}(w_n|w_1,w_2....w_{n-1})=\frac{f(w_1,w_2....w_n)}{f(w_1,w_2....w_{n-1})}\]

But again the denominator $f(w_1,w_2....w_{n-1})$ will frequently be zero,
especially for large $n$.
The backed-off estimate is a method of combating the sparse data problem.
It is defined recursively as follows:

If $f(w_1,w_2....w_{n-1})>c_1$

\[\hat{p}(w_n|w_1,w_2....w_{n-1})=\frac{f(w_1,w_2....w_n)}{f(w_1,w_2....w_{n-1})}\]

Else if $f(w_2,w_3....w_{n-1})>c_2$

\[\hat{p}(w_n|w_1,w_2....w_{n-1})=\alpha_1\times\frac{f(w_2,w_3....w_n)}{f(w_2,w_3....w_{n-1})}\]

Else if $f(w_3,w_4....w_{n-1})>c_3$

\[\hat{p}(w_n|w_1,w_2....w_{n-1})=\alpha_1\times\alpha_2\times\frac{f(w_3,w_4....w_n)}{f(w_3,w_4....w_{n-1})}\]

Else backing-off continues in the same way.

The idea here is to use MLE estimates based on lower order
n-grams if counts are not high enough to make an accurate
estimate at the current level.
The cut off frequencies ($c_1$, $c_2$....) are thresholds
determining whether to back-off or not at each level - counts
lower than $c_i$ at
stage $i$ are deemed to be too low to give an accurate estimate, so in
this case
backing-off continues. ($\alpha_1$, $\alpha_2$,....) are normalisation
constants which ensure that conditional probabilities sum to one.

Note that the estimation of $\hat{p}(w_n|w_1,w_2....w_{n-1})$ is
analogous to the estimation of $\hat{p}(1|v,n1,p,n2)$, and the above
method can therefore also be applied to the PP-attachment problem.
For example a simple method for estimation of $\hat{p}(1|v,n1,p,n2)$
would go from MLE estimates of $\hat{p}(1|v,n1,p,n2)$ to $\hat{p}(1|v,n1,p)$
to $\hat{p}(1|v,n1)$ to $\hat{p}(1|v)$ to $\hat{p}(1)$.
However a crucial difference between the two problems is
that in the n-gram task the words $w_1$ to $w_n$ are sequential,
giving a natural order in which backing off
takes place - from $\hat{p}(w_n|w_1,w_2....w_{n-1})$ to
$\hat{p}(w_n|w_2,w_3....w_{n-1})$ to
$\hat{p}(w_n|w_3,w_4....w_{n-1})$ and so on. There is no
such sequence in the PP-attachment problem,
and because of this there are four possible triples when backing off from
quadruples ($(v,n1,p)$, $(v,p,n2)$, $(n1,p,n2)$ and $(v,n1,n2)$) and six
possible pairs when backing off from triples ($(v,p)$, $(n1,p)$,
$(p,n2)$, $(v,n1)$, $(v,n2)$ and $(n1,n2)$).

A key observation in choosing between these tuples is that
the preposition is particularly important to the attachment
decision. For this reason only tuples which contained the preposition were
used in backed off estimates - this reduces the problem to a choice
between 3 triples and 3 pairs at each respective stage.
Section 6.2 describes experiments which show that
tuples containing the preposition are much better indicators of attachment.

The following method of combining the counts was found to work
best in practice:

\[\hat{p}_{triple}(1|v,n1,p,n2)=\frac{f(1,v,n1,p)+f(1,v,p,n2)+f(1,n1,p,n2)}{f(v,n1,p)+f(v,p,n2)+f(n1,p,n2)}\]

and

\[\hat{p}_{pair}(1|v,n1,p,n2)=\frac{f(1,v,p)+f(1,n1,p)+f(1,p,n2)}{f(v,p)+f(n1,p)+f(p,n2)}\]

Note that this
method effectively gives more weight to tuples with high overall
counts. Another obvious method of combination, a simple
average\footnote{eg. A simple average for triples would be defined as
\[\hat{p}_{triple}(1|v,n1,p,n2)=\frac{\frac{f(1,v,n1,p)}{f(v,n1,p)}+\frac{f(1,v,p,n2)}{f(v,p,n2)}+\frac{f(1,n1,p,n2)}{f(n1,p,n2)}}{3}\]}, gives equal weight to the
three tuples regardless of their total counts and does not perform as well.

The cut-off frequencies must then be chosen. A surprising difference
from language modeling is that a cut-off frequency of $0$ is found to
be optimum at all
stages. This effectively means however low a count is, still
use it rather than backing off a level.

\subsection{Description of the Algorithm}

The algorithm is then as follows:

\begin{enumerate}

\item  {\bf If}\footnote{At stages 1 and 2 backing off
was also continued if $\hat{p}(1|v,n1,p,n2)=0.5$. ie. the counts
were `neutral' with respect to attachment at this stage.} $f(v,n1,p,n2)>0$

	\[\hat{p}(1|v,n1,p,n2)=\frac{f(1,v,n1,p,n2)}{f(v,n1,p,n2)}\]

\item {\bf Else if} $f(v,n1,p)+f(v,p,n2)+f(n1,p,n2)>0$

\[\hat{p}(1|v,n1,p,n2)=\frac{f(1,v,n1,p)+f(1,v,p,n2)+f(1,n1,p,n2)}{f(v,n1,p)+f(v,p,n2)+f(n1,p,n2)}\]

\item {\bf Else if} $f(v,p)+f(n1,p)+f(p,n2)>0$

\[\hat{p}(1|v,n1,p,n2)=\frac{f(1,v,p)+f(1,n1,p)+f(1,p,n2)}{f(v,p)+f(n1,p)+f(p,n2)}\]

\item {\bf Else if} $f(p)>0$

	\[\hat{p}(1|v,n1,p,n2)=\frac{f(1,p)}{f(p)}\]

\item {\bf Else} $\hat{p}(1|v,n1,p,n2)=1.0$ (default is noun attachment).

\end{enumerate}

The decision is then:

If $\hat{p}(1|v,n1,p,n2)>=0.5$ choose noun attachment.

Otherwise choose verb attachment

\section{Results}

The figure below shows the results for the method on the 3097 test
sentences, also giving the total count and accuracy at each of the
backed-off stages.

\begin{center}\begin{tabular}{|c|c|c|c|} \hline
Stage&Total Number&Number Correct&Percent Correct\\ \hline \hline
Quadruples&148&134&90.5\\ \hline
Triples&764&688&90.1\\ \hline
Doubles&1965&1625&82.7\\ \hline
Singles&216&155&71.8\\ \hline
Defaults&4&4&100.0\\ \hline \hline
Totals&3097&2606&84.1\\ \hline
\end{tabular}\end{center}

\subsection{Results with Morphological Analysis}

In an effort to reduce sparse data problems the following processing
was run over both test and training data:

\begin{itemize}

\item All 4-digit numbers were replaced with the string `YEAR'.

\item All other strings of numbers (including those which had commas
or decimal points) were replaced with the token `NUM'.

\item The verb and preposition fields were converted entirely to lower
case.

\item In the n1 and n2 fields all words starting with a capital letter
followed by one or more lower case letters were replaced with `NAME'.

\item All strings `NAME-NAME' were then replaced by `NAME'.

\item All verbs were reduced to their morphological stem using the
morphological analyser described in \cite{ksze}.

\end{itemize}

These modifications are similar to those performed on the corpus used
by \cite{br}.

The result using this modified corpus was 84.5\%, an improvement
of 0.4\% on the previous result.

\begin{center}\begin{tabular}{|c|c|c|c|} \hline
Stage&Total Number&Number Correct&Percent Correct\\ \hline \hline
Quadruples&242&224&92.6\\ \hline
Triples&977&858&87.8\\ \hline
Doubles&1739&1433&82.4\\ \hline
Singles&136&99&72.8\\ \hline
Default&3&3&100.0\\ \hline \hline
Totals&3097&2617&84.5\\ \hline
\end{tabular}\end{center}

\subsection{Comparison with Other Work}

Results from \cite{rrr}, \cite{br} and the backed-off
method are shown in the table below\footnote{Results for \cite{br} with
words and classes were not available on the IBM data}.
All results are for the IBM data.
These figures should be taken in the context of the lower and upper
bounds of 72.2\%-88.2\% proposed in section 2.3.

\begin{center}\begin{tabular}{|c|c|}\hline
Method&Percentage Accuracy\\ \hline \hline
\cite{rrr} (words only)&77.7\\ \hline
\cite{rrr} (words and classes)&81.6\\ \hline
\cite{br} (words only)&81.9\\ \hline
Backed-off (no processing)&84.1\\ \hline
Backed-off (morphological processing)&84.5\\ \hline
\end{tabular}\end{center}

On the surface the method described in \cite{hr} looks
very similar to the backed-off estimate. For this reason the two
methods deserve closer comparison. Hindle and Rooth used a partial parser
to extract head nouns from a
corpus, together with a preceding verb and a following preposition,
giving a table of $(v,n1,p)$ triples. An
iterative, unsupervised method was then used to decide between noun
and verb attachment for each triple. The decision was made as
follows\footnote{This ignores refinements to
the test such as smoothing of the estimate, and a measure of the
confidence of the decision. However the measure given is at the core of
the algorithm.}:

If

\[\frac{f(n1,p)}{f(n1)}>=\frac{f(v,p)}{f(v)}\]

then choose noun attachment, else choose verb attachment.

Here $f(w,p)$ is the number of times preposition $p$ is seen attached
to word $w$ in the table, and $f(w)=\sum_p{f(w,p)}$.

If we ignore $n2$ then the IBM data is equivalent to Hindle and
Rooth's $(v,n1,p)$ triples, with the advantage of the attachment
decision being known, allowing a supervised algorithm.
The test used in \cite{hr} can then be stated
as follows in our notation:

If

\[\frac{f(1,n1,p)}{f(1,n1)}>=\frac{f(0,v,p)}{f(0,v)}\]

then choose noun attachment, else choose verb attachment.

This is effectively a comparison of the maximum likelihood estimates of
$\hat{p}(p|1,n1)$ and $\hat{p}(p|0,v)$, a different measure from the
backed-off estimate which gives $\hat{p}(1|v,p,n1)$.

The backed-off method based on just the $f(v,p)$ and $f(n1,p)$ counts
would be:

If

\[\hat{p}(1|v,n1,p)>=0.5\]

then choose noun attachment, else choose verb attachment,

where

\[\hat{p}(1|v,n1,p)=\frac{f(1,v,p)+f(1,n1,p)}{f(v,p)+f(n1,p)}\]

An experiment was implemented to investigate the difference in
performance between these two methods. The test set was restricted
to those cases where $f(1,n1)>0$, $f(0,v)>0$, and Hindle and Rooth's
method gave a definite decision. (ie. the above inequality is strictly
less-than or greater-than). This gave 1924 test cases.
Hindle and Rooth's method scored 82.1\% accuracy (1580 correct) on this set,
whereas the backed-off measure scored 86.5\% (1665 correct).

\section{A Closer Look at Backing-Off}

\subsection{Low Counts are Important}

A possible criticism of the backed-off estimate is that it uses low count
events without any
smoothing, which has been shown to be a mistake in similar
problems such as n-gram language models. In particular, quadruples and
triples seen in test data will frequently be seen only once or twice
in training data.

An experiment was made with all counts less than 5 being put to
zero,\footnote{Specifically: if for a subset  $x$ of the quadruple
$f(x)<5$, then make $f(x)=f(1,x)=f(0,x)=0$.}
effectively making the algorithm ignore low count events. In \cite{rrr}
a cut-off `between 3 and 5' is used for all events. The
training and test data were both the unprocessed, original data sets.

The results were as follows:

\begin{center}\begin{tabular}{|c|c|c|c|} \hline
Stage&Total Number&Number Correct&Percent Correct\\ \hline \hline
Quaduples&39&38&97.4\\ \hline
Triples&263&243&92.4\\ \hline
Doubles&1849&1574&85.1\\ \hline
Singles&936&666&71.2\\ \hline
Defaults&10&5&50.0\\ \hline \hline
Totals&3097&2526&81.6\\ \hline
\end{tabular}\end{center}

The decrease in accuracy from 84.1\% to 81.6\% is clear evidence for the
importance of low counts.

\subsection{Tuples with Prepositions are Better}

We have excluded tuples which do not contain a preposition from the
model. This section gives results which justify this.

The table below gives accuracies for the sub-tuples at each stage of
backing-off. The accuracy figure for a particular tuple
is obtained by modifying the
algorithm in section 4.1 to use only information from that tuple at the
appropriate stage. For example for $(v,n1,n2)$, stage 2 would be
modified to read

If $f(v,n1,n2)>0$,
\[\hat{p}(1|v,n1,p,n2)=\frac{f(1,v,n1,n2)}{f(v,n1,n2)}\]

All other
stages in the algorithm would be unchanged. The accuracy figure is
then the percentage accuracy on the test cases where the $(v,n1,n2)$
counts were used. The development set with no morphological processing
was used for these tests.

\begin{center}\begin{tabular}{|c|c||c|c||c|c|}\hline
\multicolumn{2}{|c||}{Triples}&\multicolumn{2}{|c||}{Doubles}&\multicolumn{2}{|c|}{Singles}\\ \hline
Tuple&Accuracy&Tuple&Accuracy&Tuple&Accuracy\\ \hline \hline
n1 p n2&90.9&n1 p&82.1&p&72.1\\ \hline
v p n2&90.3&v p&80.1&n1&55.7\\ \hline
v n1 p &88.2&p n2&75.9&v&52.7\\ \hline
v n1 n2 &68.4&n1 n2&65.4&n2&47.4\\ \hline
&&v n1&59.0&&\\ \hline
&&v n2&53.4&&\\ \hline
\end{tabular}\end{center}

At each stage there is a sharp difference in accuracy between tuples
with and without a preposition. Moreover, if the 14 tuples in the
above table were ranked by accuracy, the top 7 tuples would be the 7
tuples which contain a preposition.

\section{Conclusions}

The backed-off estimate scores appreciably better than other methods
which have been tested on the Wall Street Journal corpus. The accuracy
of 84.5\% is close to the human performance figure of 88\% using the 4
head words alone. A particularly surprising result is the significance
of low count events in training data. The algorithm has the additional
advantages of being conceptually simple, and computationally
inexpensive to implement.

There are a few possible improvements which may raise
performance further. Firstly, while we have shown the importance of
low-count events, some kind of smoothing may improve performance further -
this needs to be investigated. Word-classes of semantically similar
words may be used to help the sparse data problem - both \cite{rrr}
and \cite{br} report significant improvements through the use of
word-classes. Finally, more training data is almost
certain to improve results.

\end{document}